\documentclass[letterpaper]{article} 
\usepackage{aaai2027}  
\usepackage[hyphens]{url}  
\usepackage{graphicx} 
\usepackage{natbib}  
\usepackage{caption} 
\usepackage{algorithm}
\usepackage{algorithmic}

\usepackage{newfloat}
\usepackage{listings}
\DeclareCaptionStyle{ruled}{labelfont=normalfont,labelsep=colon,strut=off} 
\floatstyle{ruled}
\newfloat{listing}{tb}{lst}{}
\floatname{listing}{Listing}

\usepackage{booktabs}

\nocopyright 

\usepackage{amsmath}
\usepackage{multirow}
\usepackage{amssymb}
\title{BASeg: Boundary-Aware Remote Sensing Segmentation with Structural Penalties}
\author{
    Yuexi Song\textsuperscript{\rm 1},
    Kailai Sun\textsuperscript{\rm 2,\rm 3}\thanks{Corresponding author.},
    Zhuoyu Wang\textsuperscript{\rm 1},
    Mingyi He\textsuperscript{\rm 3},
    Paul Pu Liang\textsuperscript{\rm 3},
    Shenhao Wang\textsuperscript{\rm 4},
    Jinhua Zhao\textsuperscript{\rm 3}
}

\affiliations{
    \textsuperscript{\rm 1}National University of Singapore\\
    \textsuperscript{\rm 2}Singapore-MIT Alliance for Research and Technology\\
    \textsuperscript{\rm 3}Massachusetts Institute of Technology\\
    \textsuperscript{\rm 4}University of Florida\\
    yuexi06@u.nus.edu, skl24@mit.edu, 
    wang.zhuoyu@u.nus.edu, mingyihe@mit.edu, \\
    ppliang@mit.edu, shenhaowang@ufl.edu, jinhua@mit.edu
}

\begin{document}

\maketitle

\begin{abstract}
Semantic segmentation is a core computer vision task in the remote sensing field, accelerating advancements in urban development, agriculture, ecology, water resources, and environmental monitoring. However, recent methods usually struggle to capture fine-grained object features and boundary details. Besides, current widely used datasets often lack city morphology diversity and segmentation on generative images remains largely unexplored. To address these issues, we propose a Mahalanobis-Angle Boundary Loss (MABL) that explicitly enhances boundary and shape consistency. MABL jointly models structural importance and boundary orientation through Mahalanobis distance-based weighting and angle-aware penalty. It can be readily integrated into diverse segmentation architectures and consistently improves their accuracy. Built upon MABL, we introduce BASeg, a boundary-aware remote sensing segmentation framework with Structural Penalties. BASeg integrates a Global Visual State Space module (GSM) with a Cross-Feature Fusion module (CFM) to capture both long-range contextual dependencies and fine-grained local details. Additionally, we establish a global 10-city benchmark dataset (GCD-25k) to facilitate accurate building and road segmentation. Extensive experiments on four remote-sensing benchmarks demonstrate that BASeg consistently outperforms existing methods, achieving up to a 2.8\% improvement in mIoU while producing more accurate object boundary segmentation across diverse scenes. Moreover, integrating MABL into multiple existing segmentation architectures consistently improves performance across datasets, demonstrating its robustness and broad applicability.
\end{abstract}

\begin{links}
    \link{Code}{https://github.com/YosieSYX/BASeg}
    \link{GCD-25k Dataset}{https://www.kaggle.com/ds/8560365}
\end{links}

\section{Introduction}
\label{sec:intro}

\begin{figure*}[t]
    \centering
    \includegraphics[width=0.9\textwidth]{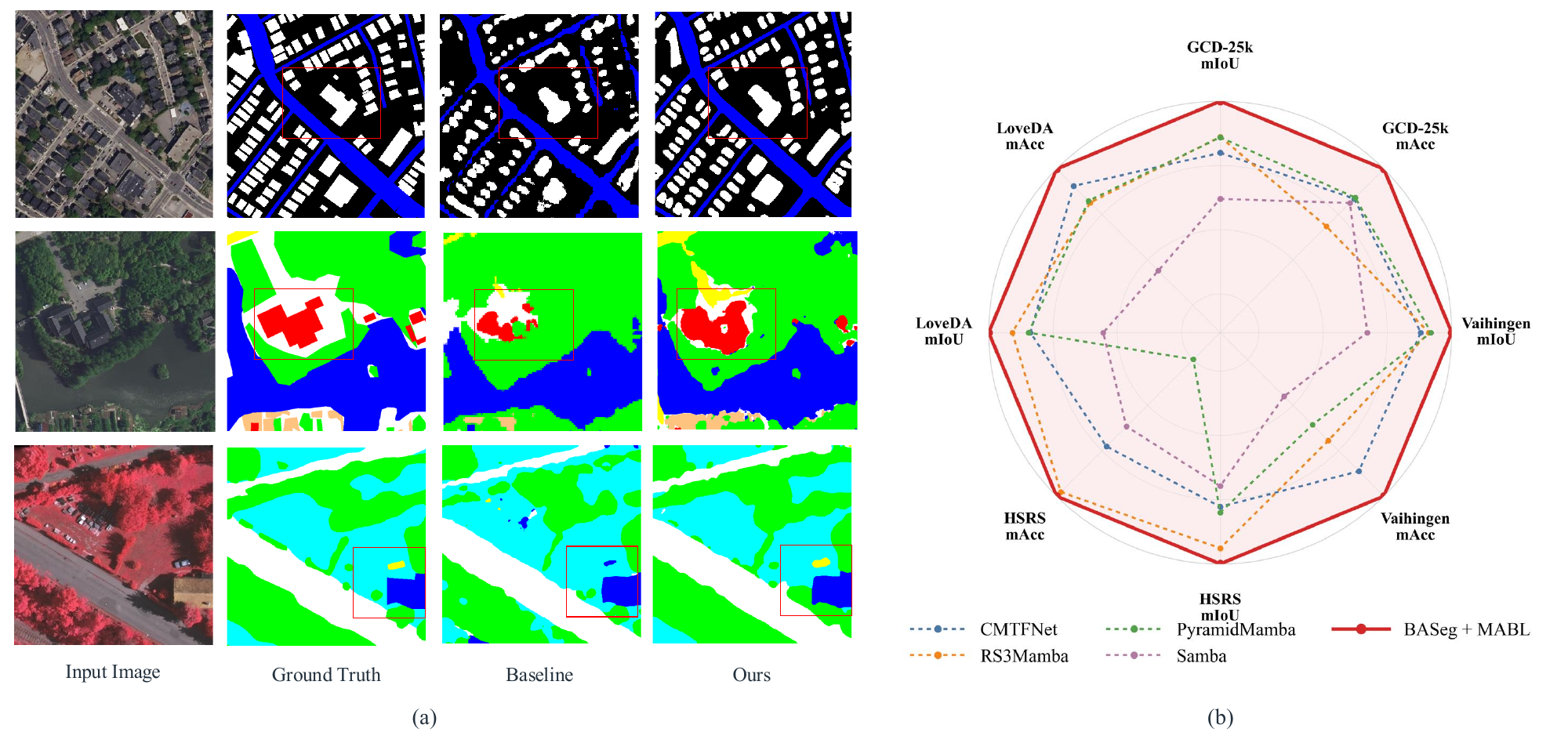}
    \caption{((a) Qualitative comparison of semantic segmentation results between BASeg and a representative baseline. Red rectangles highlight regions where BASeg produces more accurate boundaries and better separates adjacent objects. (b) Radar chart comparing mIoU and mAcc across the four benchmark datasets.}
    \label{fig:Intro}
\end{figure*}

Semantic segmentation is a fundamental computer vision task, enabling fine-grained scene understanding by assigning semantic labels to every pixel in an image \cite{semanticSegmentation}. In remote sensing, accurate semantic segmentation facilitates development in urban analytics, agriculture, ecology, water resources, and environmental monitoring \cite{li2025segearth, sense}.
Different from natural scenes, remote sensing images present unique challenges in semantic segmentation, including significant scale variations of foreground objects, extremely large image sizes, complex backgrounds \cite{huang2025multi}, as well as tiny
and densely distributed objects\cite{qiao2025sam}. 

Recent advances in semantic segmentation for Earth observation imagery have explored both optimization and architectural improvements. On the optimization side, numerous loss functions have been proposed to enhance boundary segmentation. Distance-map-based loss \cite{distance_loss} and Hausdorff-distance loss \cite{HausdoffLoss} supervise boundary localization through distance-based objectives. Active contour loss \cite{ContourLoss} encourages contour alignment, while Active Boundary Loss (ABL) \cite{wang2022active} enforces boundary alignment through directional supervision. 
More recently, methods such as InverseForm \cite{inverseForm} and SAM-guided boundary supervision \cite{SAM_RS} exploit auxiliary networks or pretrained foundation models to learn richer boundary representations for segmentation.
Architecturally, existing works have explored CNNs  and Transformers to improve segmentation performance in complex urban environments. Recent state-space models \cite{Mamba, RS3Mamba} have been explored to model global context, and attention mechanisms have been used to extract finer local structures. Besides, large pretrained vision models, such as SAM \cite{SAM} and DINOv2 \cite{luo2025domain}, have been leveraged for their strong feature representations to improve segmentation performance in remote sensing tasks \cite{UFMF, SAM_RS}.  Based on SAM or diffusion models,  some studies utilize edge prior information  \cite{gan2025prior,wang2026conditional} to guide the effective semantic feature extraction. 

Despite recent advancements, existing models still struggle to accurately segment object boundaries \cite{huang2025multi}, often failing to distinguish adjacent objects and producing blurred edges. For example, in Figure \ref{fig:Intro} (a), the predicted building contours appear irregular, deviating from the expected clean, rectilinear geometry and resulting in wavy or uneven shapes. What's more, most of these existing segmentation loss functions rely heavily on Euclidean distance and pixel-wise boundary predictions, paying insufficient attention to boundary shape and directional consistency. 

To address these research gaps, we propose BASeg, a boundary-aware semantic segmentation framework that combines global semantic modeling, local spatial encoding, cross-feature fusion, and structure-aware supervision. We introduce a general Mahalanobis-Angle Boundary loss, which comprises two components: an angle loss that penalizes orientation discrepancies between predicted and ground-truth boundaries to enforce boundary alignment and a Mahalanobis distance-based boundary loss that captures structural covariance and shape priors to better preserve object geometry. Importantly, MABL is architecture-agnostic and can be incorporated into other semantic segmentation frameworks as a general training objective. Experiments demonstrate that the proposed loss functions achieve better performance with lightweight computation, compared to existing loss functions. Building on MABL, we propose BASeg, a dual-branch boundary-aware framework that jointly captures global semantic context and fine-grained spatial details. BASeg employs a Global Visual State-Space Module (GSM) to model long-range dependencies from frozen DINOv3 features through coordinate-enhanced 2D selective scanning. A Cross-Feature Fusion Module (CFM) then integrates global semantic context with multiscale local details, followed by a coarse-to-fine decoder that progressively refines object structures and boundaries.

To summarize, our contributions are as follows:  

(1) Introduced the general Mahalanobis-Angle Boundary Loss (MABL), which combines Mahalanobis-based spatial reweighting with boundary-angle consistency. The Mahalanobis-distance loss promotes structural and geometric coherence, while the angle loss penalizes orientation discrepancies between predicted and ground-truth boundaries, producing sharper and better segmentation contours.

(2) Designed a multi-scale feature extraction architecture integrating GSM and CFM to effectively combine global contextual information with fine-grained local details for accurate semantic segmentation.

(3) Established a large-scale benchmark dataset, GCD-25k, comprising 25,000 satellite images collected from 10 major global cities for fine-grained road and building segmentation. 

(4) Experimental results on four diverse datasets demonstrate that BASeg outperforms state-of-the-art methods, achieving cleaner boundaries and higher IoU and accuracy across diverse urban scenes. Ablation studies further validate the effectiveness of the proposed components. 

\section{Related Works}
\subsection{Semantic Segmentation for Remote Sensing}

Semantic segmentation for remote sensing images has been extensively studied, with most models falling into two broad categories: CNN-based\cite{ABCNet, CMTFNet} and Transformer-based\cite{UNetformer,Bi_You_Gevers_2024} approaches. More recently, dynamic state space models (SSMs) \cite{Mamba} have been introduced to efficiently capture long-range spatial dependencies, while attention mechanisms are integrated to enhance local feature representation. Recent developments in large-scale pretrained models have further advanced remote sensing segmentation. For example, the Segment Anything Model (SAM) \cite{SAM} has demonstrated strong adaptability across tasks, and several studies \cite{UFMF, SAM_RS} leverage its feature representations for remote sensing. Some studies utilize DINOv2 \cite{luo2025domain} to enhance segmentation networks’ robustness across various unseen target domains for domain generalizable remote sensing semantic segmentation \cite{gong2025crossearth}. Instead of spatial-domain features, some recent studies \cite{10847802,gao2025adaptive} focus on the feature extraction and enhancement in the frequency domain, highlighting the high-frequency details (e.g., textures of buildings and roads).

Building on these advances, we adopt frozen DINOv3 \cite{simeoni2025dinov3}, a more recent visual foundation model that produces more robust and detailed semantic embeddings for feature extraction. We further combine state space models for global features with multi-head self-attention for local features through cross-attention mechanisms to effectively fuse global and local information.

\subsection{Segmentation Loss}
Loss functions play a crucial role in the training of segmentation models. Among them, cross-entropy loss is a widely used pixel-wise measure for evaluating classification performance. However, it sometimes struggles to accurately capture boundary discrepancies. To address this, numerous boundary-based loss functions have been proposed to enhance segmentation performance. The distance map loss \cite{distance_loss} incorporates distance maps into the cross-entropy formulation to emphasize pixels farther from the correct boundaries. Hausdorff distance \cite{HausdoffLoss} computes the distance between predicted and ground-truth boundaries, explicitly penalizing spatial misalignment. Active contour loss \cite{ContourLoss} encourages smooth boundaries by minimizing gradient magnitudes along the predicted contour. Region-wise boundary loss \cite{RWLoss} applies a larger weight to pixels on the boundary. However, these methods rely mainly on Euclidean distance and focus on pixel-wise boundary prediction, without considering boundary structure and directional consistency. To address this limitation, we incorporate a Mahalanobis distance-based term to account for better spatial orientation. 
We also propose the angle loss function, which penalizes differences in orientation between predicted and ground-truth boundaries, thereby encouraging structurally consistent and straight boundaries. 

\section{Methods}
\subsection{Mahalanobis-Angle Boundary Loss Function}

We introduce a boundary-aware optimization objective designed to improve the geometric fidelity of semantic segmentation predictions. Unlike standard pixel-wise losses that treat boundary and interior pixels uniformly, the proposed objective explicitly models two complementary aspects of object boundaries: local contour orientation and instance-level shape geometry. The overall loss is formulated as
\begin{equation}
\mathcal{L}
=
\alpha \mathcal{L}_{\mathrm{angle}}
+
\beta \mathcal{L}_{\mathrm{boundary}},
\end{equation}
where $\mathcal{L}_{\mathrm{angle}}$ enforces orientation consistency between predicted and ground truth boundaries, and $\mathcal{L}_{\mathrm{boundary}}$ applies a Mahalanobis distance-based weighting to emphasize geometrically informative boundary regions. The coefficients $\alpha$ and $\beta$ balance the contributions of the two terms.

\subsubsection{Angle Loss}

Accurate segmentation of high-resolution remote sensing images requires not only pixel-wise class correctness, but also precise boundary alignment. We therefore introduce an angle loss that explicitly penalizes orientation disagreement between the predicted probability map and the ground truth mask at boundary pixels.

Let $G_k \in \{0,1\}^{H \times W}$ denote the binary ground truth mask for class $k$, and let $P_k \in [0,1]^{H \times W}$ denote the corresponding softmax probability map. We first extract the set of ground truth boundary pixels $T_k$ from $G_k$ using the Canny edge detector. For each boundary pixel $(i,j) \in T_k$, local gradients are computed using forward finite differences:
\begin{equation}
\nabla_x G_k(i,j) = G_k(i,j+1) - G_k(i,j)
\end{equation}
\begin{equation}
\nabla_y G_k(i,j) = G_k(i+1,j) - G_k(i,j),
\end{equation}
and similarly for $P_k$. The local boundary orientations are then defined as
\begin{equation}
\theta^g_{i,j}
=
\operatorname{atan2}
\left(
\nabla_y G_k(i,j),
\nabla_x G_k(i,j)
\right),
\end{equation}
\begin{equation}
\theta^p_{i,j}
=
\operatorname{atan2}
\left(
\nabla_y P_k(i,j),
\nabla_x P_k(i,j)
\right).
\end{equation}

Since semantic boundaries are unoriented, directions that differ by $\pi$ represent the same contour orientation. We therefore measure the angular discrepancy as
\begin{equation}
\Delta \theta_{i,j} = \min \left( |\theta^g_{i,j} - \theta^p_{i,j}|, \pi - |\theta^g_{i,j} - \theta^p_{i,j}| \right).
\end{equation}
The final angle loss is averaged independently over classes:
\begin{equation}
\mathcal{L}_{\mathrm{angle}}
=
\frac{1}{K}
\sum_{k=1}^{K}
\frac{1}{|T_k|}
\sum_{(i,j)\in T_k}
\left(
1 -
\cos(\Delta \theta_{i,j})
\right).
\end{equation}
This per-class formulation prevents classes with many boundary pixels from dominating the loss and gives each semantic category equal influence.

\subsubsection{Mahalanobis Distance-Based Boundary Loss}

\begin{figure}[t]  
    \centering
    \includegraphics[width=0.9\columnwidth]{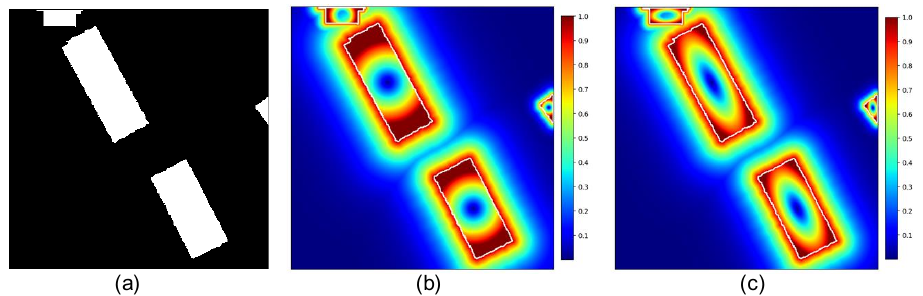} 
    \caption{(a) Ground truth segmentation, (b) Euclidean distance map and (c) Mahalanobis distance map. The Mahalanobis prior better adapts to each instance's orientation and geometry.}
    \label{fig:Mahalanobis}
\end{figure}

The second term introduces a Mahalanobis-distance-based structural prior that adaptively weights pixels according to their spatial distribution within each object region. While Euclidean distance treats all spatial directions equally, the Mahalanobis distance incorporates the covariance of each connected region, allowing the loss to adapt to object elongation, scale, and orientation. Figure \ref{fig:Mahalanobis} illustrates the Mahalanobis-based shape prior better captures the structural characteristics and orientation of each building instance compared to the Euclidean-based prior.

For each class $k$, we first construct a binary mask from the ground truth label map and extract connected components using 8-neighborhood connectivity. Each connected component is treated as an individual region $r$, denoted by $\Omega_{r,k}$. This allows the shape prior to be computed independently for each object instance. The Mahalanobis prior is computed only for foreground regions. Background pixels are excluded because the background typically forms a large, irregular connected component whose covariance does not represent a meaningful object geometry.

For each connected region $r$ of class $k$, let $\Omega_{r,k}$ denote its pixel set. We compute its centroid $\mu_{r,k}$ and covariance matrix $\Sigma_{r,k}$:
\begin{equation}
\Sigma_{r,k}
=
\begin{bmatrix}
\mathrm{Var}(x) & \mathrm{Cov}(x,y) \\
\mathrm{Cov}(y,x) & \mathrm{Var}(y)
\end{bmatrix}.
\end{equation}

To ensure numerical stability for small or degenerate connected components, we regularize the covariance matrix as
\begin{equation}
\widetilde{\Sigma}_{r,k}
=
\Sigma_{r,k}+\epsilon I,
\end{equation}
where $I$ is the identity matrix and $\epsilon=10^{-5}$. The regularized matrix $\widetilde{\Sigma}_{r,k}$ is used in the Mahalanobis-distance computation. For single-pixel components, where a meaningful covariance cannot be estimated, we set the corresponding Mahalanobis prior to zero. For two-pixel components, we use $\widetilde{\Sigma}_{r,k}=\epsilon I$ to avoid singular covariance inversion.

For a pixel coordinate $n \in \Omega_{r,k}$, the Mahalanobis distance is
\begin{equation}
d_{r,k}(n)
=
\sqrt{
(n-\mu_{r,k})^\top
\widetilde{\Sigma}_{r,k}^{-1}
(n-\mu_{r,k})
}.
\end{equation}
The distance is normalized within each connected region:
\begin{equation}
\Phi_{r,k}(n)
=
\frac{
d_{r,k}(n)
}{
\max_{m \in \Omega_{r,k}} d_{r,k}(m)
}.
\end{equation}

The Mahalanobis prior is then used to weight the cross-entropy loss:
\begin{equation}
\mathcal{L}_{\mathrm{boundary}}
=
-
\frac{1}{|\Omega|}
\sum_{n \in \Omega}
\left(
1 + \gamma \Phi(n)
\right)
\log p_{n,y_n},
\end{equation}
where $\Omega$ is the set of valid pixels, $y_n$ is the ground truth class label of pixel $n$, $p_{n,y_n}$ is the predicted probability for the correct class, $\Phi(n)$ is the normalized Mahalanobis prior of the connected region containing $n$, and $\gamma$ controls the strength of the boundary-aware weighting.

\subsection{Network Architecture}
\begin{figure}[t]  
    \centering
    \includegraphics[width=\columnwidth]{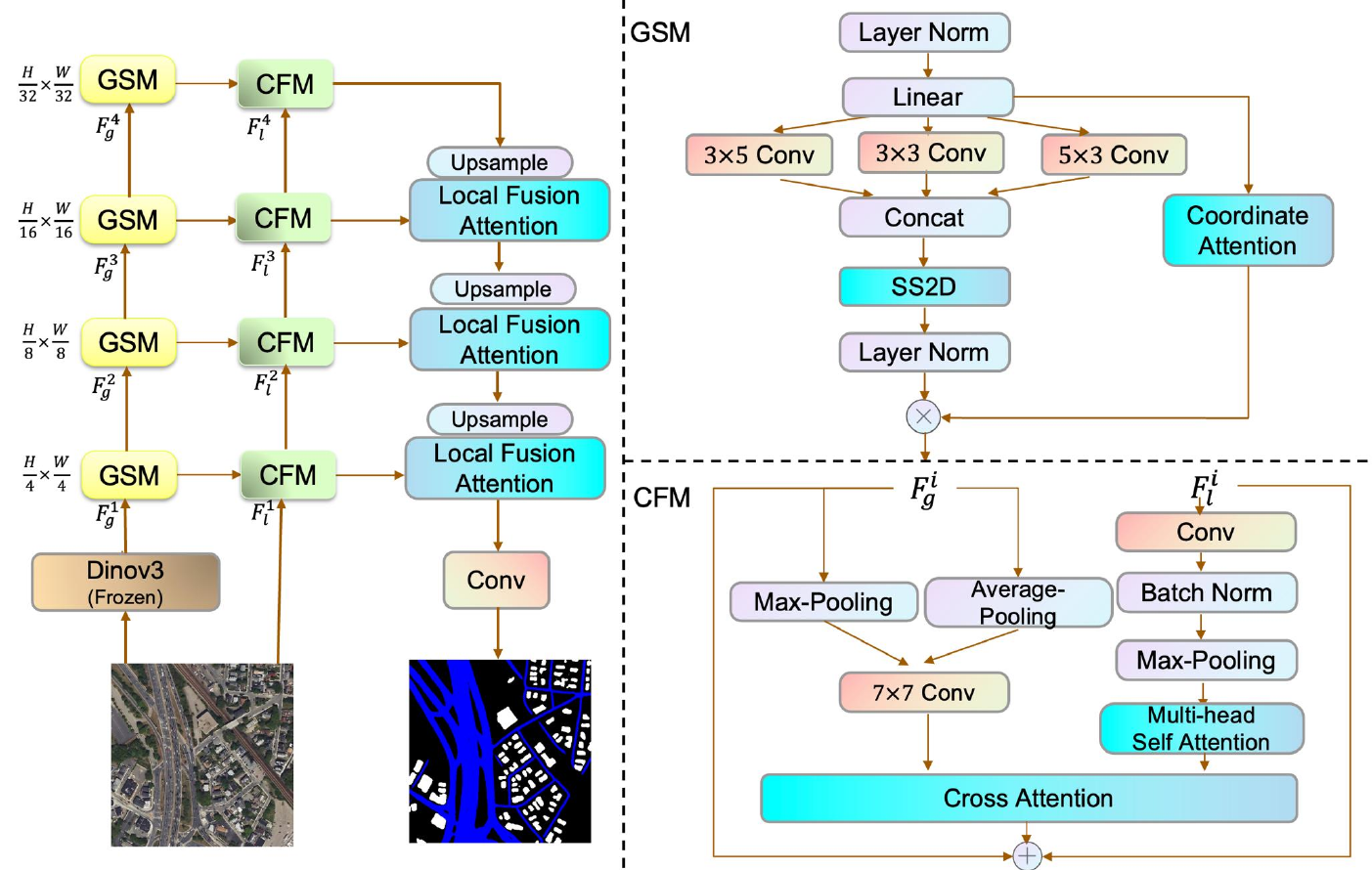} 
    \caption{Overall Architecture of BASeg.}
    \label{fig:BASeg}
\end{figure}
As illustrated in Figure~\ref{fig:BASeg}, BASeg comprises two principal components: the Global Visual State-Space Module (GSM) and the Cross-Feature Fusion Module (CFM). First, a frozen DINOv3 feature extractor~\cite{simeoni2025dinov3} transforms the input image into high-level semantic tokens. GSM projects these tokens into a hierarchical representation and processes them using coordinate-enhanced two-dimensional selective scanning to capture multiscale global context and structural information. At each encoder stage, CFM integrates the corresponding GSM extracted features through spatial attention, window-based self-attention, and cross-attention. The decoder follows a coarse-to-fine reconstruction strategy. Beginning with the deepest fused feature, each decoding stage upsamples the current representation and combines it with a higher-resolution fused encoder feature through learnable weighted fusion. Window-based attention and feature-refinement operations further enhance the reconstructed features before the segmentation head restores the prediction to the original image resolution. This dual-stream design enables BASeg to combine long-range semantic context with fine-grained spatial details.

\subsubsection{Global Visual State-Space Module (GSM)}

GSM is organized as a four-stage hierarchical encoder whose spatial resolution is progressively reduced while its feature dimensionality increases. Feature vectors from DINOv3 are first projected to the GSM embedding dimension, reshaped into a two-dimensional feature map, and resized to initialize the hierarchical encoder. Within each stage, the features are first processed through one rectangular convolution module with kernel sizes $3 \times 5$, $3 \times 3$ and $5 \times 3$, specifically designed to capture structural characteristics of rectangular objects commonly found in remote sensing imagery, such as buildings, cars, and courts. Their outputs are concatenated and processed by the 2D Selective Scan module~\cite{vmamba}, which models long-range dependencies along multiple spatial directions. In parallel, coordinate attention~\cite{coordinateAttention} encodes position-sensitive responses along the horizontal and vertical axes. The coordinate-aware features then gate the selective-scan output, enhancing the representation of both global context and directional structure.

\subsubsection{Cross-feature Fusion Module (CFM)}
The CFM comprises two parallel branches: the global branch and the local branch. The global branch receives the feature map $F^{i}_g$ from the corresponding GSM stage and refines it using Spatial Attention to capture global contextual and structural information. Specifically, max and average pooling responses are concatenated and combined through a $7 \times 7$ convolution. Simultaneously, the local branch applies multi-head self-attention to extract detailed, fine-grained features $F^{i}_l$.
Then, to enhance global and local feature fusion, we employ a cross-attention module, where the local branch serves as the query vector, and the global branch provides the key and value vectors. This allows the local features to selectively integrate relevant information from the global context. Finally, the output from cross attention is combined with the original local and global features, forming a residual connection that preserves both local details and global context.

\section{Experiments}
We conduct extensive experiments to answer the following research questions:

\begin{itemize}
    \item RQ1: How does the proposed BASeg framework perform compared to state-of-the-art remote sensing semantic segmentation methods across diverse benchmark datasets?
    
    \item RQ2: Can the proposed MABL be effectively applied and generalized to other existing remote sensing semantic segmentation architectures?    
    \item RQ3: How does MABL compare with existing boundary-aware losses and how do its individual components contribute to performance?
    
\end{itemize}
\subsection{Datasets}
\textbf{Proposed Dataset}

\textit{1) GCD-25k: }
There are several publicly available datasets for remote sensing semantic segmentation, but many of them are limited in scale and geographic diversity. 
Given the lack of variation in the urban morphology of existing datasets, we establish a new dataset (\textbf{GCD-25k}), comprising 25,000 satellite images from 10 global cities and it is annotated with three classes: background, building, and road. To ensure geographic diversity, the dataset covers 10 cities across Asia (Singapore, Tokyo), Africa (Abuja, Kinshasa), the Americas (Boston, San Francisco, São Paulo), Europe (Barcelona, Paris), and Australia (Perth), with each city contributing 2500 images. Satellite imagery is obtained from Mapbox, 
while road networks and building footprints are derived from OpenStreetMap, where the vector data are spatially aligned with the satellite imagery and then rasterized to generate pixel-level segmentation masks. Each image has a resolution of 512 × 512 pixels, covering 400 m × 400 m of ground area. Image patches are sampled within each city boundary to ensure spatial coverage and diversity. We have divided it into 2 subsets: training (20000), testing (5000). For each city, 2,000 images are assigned to the training set and the remaining 500 to the test set, ensuring balanced geographic representation across both subsets. More details can be found in the supplementary material.

\textbf{Additional Evaluation Datasets}

\textit{2) HSRS Dataset: }
Existing semantic segmentation on generative images by generative AI models remains largely unexplored. Thus, we use a Hybrid-Synthetic-Real-Segmentation dataset (HSRS) \cite{11223686} containing 3,000 manually annotated images (extended version). The dataset combines real samples with synthetic images generated by the fine-tuned SDXL model, where the generated samples are further processed by a data-cleaning pipeline to select high-quality images. The dataset contains 9 categories: background, road, building, vegetation, car, water, field, court, and square.

\textit{3) LoveDA Urban: } 
LoveDA~\cite{wang2021loveda} contains 1,156 training images and 677 testing images collected from three cities in China in both the urban and rural domains.

\textit{4) ISPRS Vaihingen: }
The ISPRS Vaihingen dataset~\cite{ISPRS} contains 33 high-resolution orthophotos tiles (2500 × 2000 pixels). We further cropped into 512 × 512 patches (stride size: 256 pixels), yielding 1,294 training and 442 testing images. 
\subsection{Implementation details}
We initialized VMamba \cite{vmamba} with ImageNet-pretrained weights and fine-tuned it during training, while keeping the pretrained DINO feature extractor frozen throughout. We trained the models using the SGD optimiser for 50 epochs with a batch size of 32. We evaluated the segmentation performance using the intersection over union (IoU) and accuracy (Acc) of each class, and the overall mean intersection over union 
(mIoU) as well as mean accuracy (mAcc). All results are reported as percentages. The learning rate is set to 0.001 with weight decay of 0.01. All models were trained on 4 NVIDIA H100 GPUs. For the loss function, the weighting coefficients are set to $\alpha = 1$ and $\beta = 0.5$, and the Mahalanobis shape prior scaling factor is set to $\gamma = 0.5$. 

\begin{figure*}[t]  
    \centering
    \includegraphics[width=0.9\textwidth]{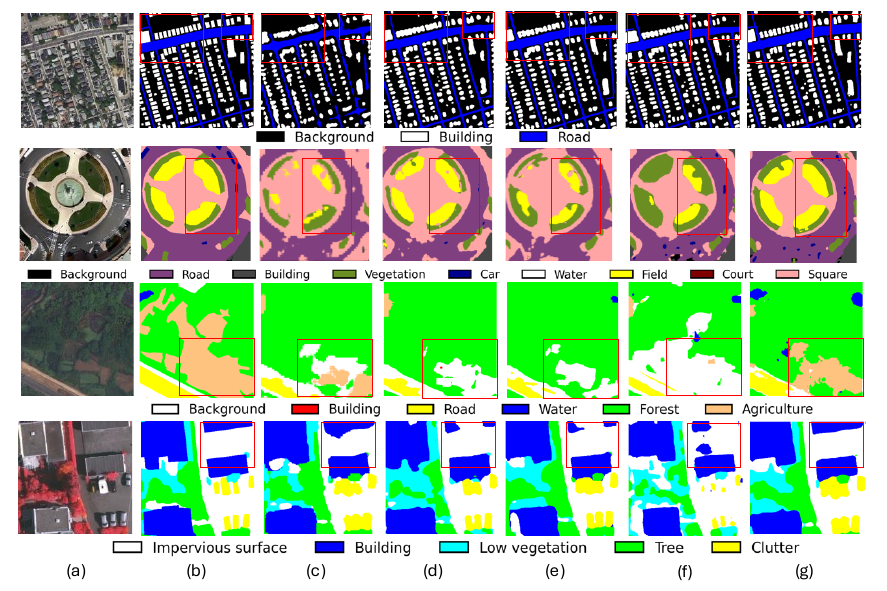} 
    \caption{Qualitative comparison and analysis on four datasets: (a) Original images, (b) Ground truth, (c) ABCNet, (d) UNetFormer, (e) CMTFNet, (f) RS3Mamba, and (g) the proposed BASeg.}
    \label{fig:Qualitative}
\end{figure*}

\subsection{Main Results}
We thoroughly evaluate our method on four datasets: our proposed GCD-25k dataset, HSRS dataset, and two widely used public benchmarks, LoveDA and ISPRS Vaihingen. We compare our approach against state-of-the-art remote sensing semantic segmentation models, including RS3Mamba~\cite{RS3Mamba}; SAM\_RS\cite{SAM_RS} combined with ABCNet~\cite{ABCNet}, and Unetformer~\cite{UNetformer}; CMTFNet~\cite{CMTFNet}, PyramidMamba~\cite{wang2024pyramidmamba} and Samba~\cite{zhu2024samba}.

As shown in Table~\ref{tab:main_results}, BASeg + MABL achieves the best overall performance across all four benchmarks. On GCD-25k, it achieves the highest performance across all classes, obtaining an mIoU of 68.86\%, outperforming the previous best method by 2.77\%. On the Vaihingen benchmark, the proposed method improves Building IoU to 88.25\% and Car IoU to 57.64\%, highlighting its effectiveness in preserving object geometry and accurately delineating small objects. On HSRS, BASeg + MABL exceeds the strongest baseline by 1.22\%. Similarly, on LoveDA, it outperforms all competing methods with 47.53\% mIoU and 68.09\% mAcc. These consistent improvements across four datasets demonstrate that BASeg effectively captures both global semantic context and fine-grained boundary information, resulting in improved segmentation quality under diverse remote sensing scenarios.

Table~\ref{tab:main_results} also demonstrates that the proposed Mahalanobis-Angle Boundary Loss (MABL) is architecture-agnostic and consistently improves the performance of existing segmentation networks. When integrated into PyramidMamba, MABL improves mIoU on all four datasets, yielding gains of 1.01\%, 2.21\%, 2.25\%, and 1.87\% on GCD-25k, Vaihingen, HSRS, and LoveDA, respectively. Similarly, incorporating MABL into Samba consistently improves segmentation performance across all benchmarks. It also improves categories with well-defined geometric structures. For example, on the Vaihingen dataset, Car IoU increases by 6.16\% for PyramidMamba and 7.90\% for Samba. MABL also provides substantial improvements for more challenging categories, improving Tree IoU by 0.32\% and 1.16\% on PyramidMamba and Samba respectively. These consistent improvements across different architectures and semantic categories demonstrate that MABL provides complementary geometric supervision. Moreover, MABL can be seamlessly integrated into diverse segmentation architectures without requiring any network modifications.

\begin{table*}[t]
\centering
\caption{Quantitative comparison on GCD-25k, Vaihingen, HSRS, and LoveDA.
Best results are shown in bold. The $\triangle$ rows report the performance gain over the corresponding baseline.}
\label{tab:main_results}

\resizebox{\textwidth}{!}{%
\begin{tabular}{ll*{16}{c}}
\toprule

\multirow{2}{*}{Method}
& \multirow{2}{*}{Backbone}
& \multicolumn{4}{c}{GCD-25k}
& \multicolumn{6}{c}{Vaihingen}
& \multicolumn{2}{c}{HSRS}
& \multicolumn{2}{c}{LoveDA} \\

\cmidrule(lr){3-6}
\cmidrule(lr){7-12}
\cmidrule(lr){13-14}
\cmidrule(lr){15-16}

& & Bkg & Bld & Rd & mIoU
& Imp & LV & Bld & Tree & Car & mIoU
& mIoU & mAcc
& mIoU & mAcc \\

\midrule

ABCNet
& ResNet
& 73.08 & 52.57 & 41.48 & 55.71
& 73.27 & 59.85 & 80.59 & 70.92 & 51.45 & 67.22
& 62.48 & 76.46
& 44.35 & 62.54\\

UNetFormer
& ResNet
& 72.90 & 52.70 & 50.56 & 58.72
& 73.79 & 60.23 & 81.16 & 71.93 & 54.65 & 68.35
& 63.76 & 76.62
& 41.90 & 64.32\\

RS3Mamba
& Mamba+ResNet
& 77.69 & 62.68 & 57.89 & 66.09
& 77.90 & 62.28 & 81.00 & \textbf{77.37} & 51.32 & 70.77
& 68.19 & 80.31
& 45.70 & 64.39\\

\midrule


CMTFNet
& ResNet
& 73.21 & 54.76 & 45.97 & 57.98
& 78.24 & 59.80 & 85.25 & 72.54 & 50.35 & 69.24
& 65.53 & 74.11
& 40.53 & 58.22\\

CMTFNet+MABL
& ResNet
& 76.55 & 61.02 & 56.97 & 64.85
& 77.08 & 64.76 & 85.99 & 73.65 & 54.86 & 71.27
& 66.04 & 78.03
& 44.97 & 63.46\\

$\triangle$
&
& +3.34 & +6.26 & +11.00 & +6.87
& -1.16 & +4.96 & +0.74 & +1.11 & +4.51 & +2.03
& +0.51 & +3.92
& +4.44 & +5.24\\
\midrule 

PyMamba
& Swin-base
& 79.47 & 60.03 & 57.88 & 65.79
& 76.26 & 64.72 & 86.76 & 75.18 & 45.21 & 69.63
& 65.41 & 65.72
& 44.44 & 64.58\\

PyMamba+MABL
& Swin-base
& 80.63 & 61.49 & 58.27 & 66.80
& 78.75 & 66.72 & 86.84 & 75.50 & 51.37 & 71.84
& 67.66 & 77.07
& 46.31 & 67.48\\

$\triangle$
&
& +1.16 & +1.46 & +0.39 & +1.01
& +2.49 & +2.00 & +0.08 & +0.32 & +6.16 & +2.21
& +2.25 & +11.35
& +1.87 & +2.90\\

\midrule

Samba
& Samba
& 67.54 & 59.47 & 56.79 & 61.27
& 76.29 & 63.42 & 83.93 & 72.7 & 33.61 & 65.99
& 63.35 & 73.09
& 38.63 & 56.90\\

Samba+MABL
& Samba
& 71.94 & 61.69 & 59.55 & 64.39
& 76.65 & 61.59 & 84.08 & 73.86 & 41.51 & 68.74
& 64.04 & 74.64
& 41.01 & 59.49\\

$\triangle$
&
& +4.40 & +2.22 & +2.76 & +3.12
& +0.36 & -1.83 & +0.15 & +1.16 & +7.90 & +2.75
& +0.69 & +1.55
& +2.38 & +2.59\\

\midrule

\textbf{BASeg+MABL}
& \textbf{Mamba+DINOv3}
& \textbf{81.24}
& \textbf{63.87}
& \textbf{61.48}
& \textbf{68.86}
& \textbf{79.12}
& \textbf{66.46}
& \textbf{88.25}
& 73.92
& \textbf{57.64}
& \textbf{73.08}
& \textbf{69.41}
& \textbf{80.77}
& \textbf{47.53}
& \textbf{68.09}\\

\bottomrule
\end{tabular}%
}
\end{table*}

\subsection{Ablation Study}
\subsubsection{Analysis of the Proposed Loss Function}
To evaluate the contribution of each component of our loss function MABL, we conducted ablation studies with BASeg on the GCD-25k and Vaihingen dataset, as detailed in Table \ref{tab:loss_ablation}. 

The angle consistency loss enforces local directional consistency along object boundaries and is particularly useful for man-made objects with well-defined geometric structures. Compared with using the Mahalanobis loss alone, incorporating the angle consistency loss improves the Building IoU in Vaihingen by 3\% and substantially boosts the Car IoU from 46.28\% to 57.64\%.
The Mahalanobis boundary loss complements the angle consistency loss by adapting the boundary supervision to the local geometric distribution of each object. On GCD-25k, it consistently improves the segmentation of the Building and Road categories, resulting in an overall 0.6\% increase in mAcc. On the more diverse Vaihingen dataset, it is particularly effective for geometrically irregular categories, improving the Low Vegetation and Tree IoUs by 4.8\% and 2.3\%, respectively.
These results demonstrate that the two losses capture complementary geometric cues, jointly improving both boundary localization and structural coherence, leading to the highest overall mIoU and mAcc on both datasets.
\begin{table}[t]
\centering
\caption{Ablation study of the proposed loss functions on the GCD-25k and Vaihingen datasets.}
\label{tab:loss_ablation}

\scriptsize 
\setlength{\tabcolsep}{3pt} 

\begin{tabular}{c|l|cc|cc|cc}
\toprule
\multirow{2}{*}{Dataset} &
\multirow{2}{*}{Class}
& \multicolumn{2}{c|}{Only AngL}
& \multicolumn{2}{c|}{Only MaL}
& \multicolumn{2}{c}{Combined} \\
\cline{3-8}

&
& IoU & Acc
& IoU & Acc
& IoU & Acc \\
\midrule

\multirow{4}{*}{GCD-25k}

& Building
& 63.66 & 77.41
& 63.87 & 77.26
& \textbf{63.87} & \textbf{78.00} \\

& Road
& 60.40 & 74.70
& 60.20 & 76.34
& \textbf{61.48} & \textbf{77.84} \\

& Background
& 81.21 & \textbf{89.87}
& \textbf{81.36} & 89.88
& 81.24 & 89.36 \\

\cline{2-8}

& mIoU / mAcc
& 68.42 & 80.66
& 68.48 & 81.16
& \textbf{68.86} & \textbf{81.73} \\

\midrule

\multirow{6}{*}{Vaihingen}

& Impervious
& 75.40 & 85.49
& 73.66 & 83.81
& \textbf{79.72} & \textbf{87.00} \\

& Low Vegetation
& 61.63 & 73.53
& 61.55 & 75.35
& \textbf{66.46} & \textbf{77.28} \\

& Building
& 85.50 & 93.39
& 85.21 & 94.12
& \textbf{88.25} & \textbf{94.42} \\

& Tree
& 71.64 & \textbf{84.25}
& 72.03 & 82.54
& \textbf{73.92} & \textbf{84.82} \\

& Car
& 44.12 & \textbf{93.61}
& 46.28 & 92.72
& \textbf{57.64} & 88.83 \\

\cline{2-8}

& mIoU / mAcc
& 67.66 & 86.05
& 67.75 & 85.71
& \textbf{73.08} & \textbf{86.47} \\

\bottomrule
\end{tabular}
\end{table}

\subsubsection{Analysis of the Proposed Architecture}
To validate the effectiveness of each component of the proposed architecture, we perform an ablation study on the Vaihingen dataset, with additional results on GCD-25k, LoveDA, and HSRS provided in the supplementary material.
As shown in Table~\ref{tab:vaihingen_model}, introducing the Rectangular Convolution Module (RCM) improves the overall mIoU by 8.5\%, with particularly large gains on structure-aware classes such as building and car, whose IoUs increase by 5.6\% and 19.1\%, respectively. Removing Coordinate Attention results in a 3.5\% drop in mIoU, demonstrating the benefit of coordinate-aware feature modeling. We further compare the proposed Cross Fusion Module (CFM) with the CCM module adopted in RS3Mamba~\cite{RS3Mamba}. Replacing CFM with CCM decreases the overall mIoU from 73.1\% to 69.5\%, confirming the effectiveness of our feature fusion strategy. Overall, GSM and CFM provide complementary benefits, jointly yielding the best segmentation performance.

\begin{table}[!t]
    \centering
    \caption{Ablation study of GSM and CFM architectures on the Vaihingen dataset. (CCM refers to the architecture used in RS3Mamba~\cite{RS3Mamba}).}
    \label{tab:vaihingen_model}
    
    \scriptsize 
    \setlength{\tabcolsep}{3pt} 
    
    \begin{tabular}{l|cc|cc|cc|cc}
        \toprule
        GSM & \multicolumn{2}{c|}{No RCM} & \multicolumn{2}{c|}{No Coord. Attn.} & \multicolumn{2}{c|}{$\checkmark$} & \multicolumn{2}{c}{$\checkmark$} \\
        \hline
        CFM & \multicolumn{2}{c|}{$\checkmark$} & \multicolumn{2}{c|}{$\checkmark$} & \multicolumn{2}{c|}{CCM} & \multicolumn{2}{c}{$\checkmark$} \\
        \hline
        Class & IoU & Acc & IoU & Acc & IoU & Acc & IoU & Acc \\
        \hline
        Impervious      & 72.65 & 80.86 & 74.65 & 84.19 & 74.93 & 83.81 & \textbf{79.12} & \textbf{87.00} \\
        Low Vegetation  & 57.66 & 73.94 & 62.14 & 77.06 & 62.37 & 76.25 & \textbf{66.46} & \textbf{77.28} \\
        Building        & 82.66 & 93.30 & 85.47 & 93.29 & 85.66 & 94.41 & \textbf{88.25} & \textbf{94.42} \\
        Tree            & 71.49 & 81.45 & 72.18 & 82.66 & 72.36 & 83.49 & \textbf{73.92} & \textbf{84.82} \\
        Car             & 38.57 & 90.48 & 52.80 & 92.89 & 52.40 & 92.52 & \textbf{57.64} & 88.83 \\
        \hline
        mIoU / mAcc     & 64.61 & 84.01 & 69.45 & 86.02 & 69.54 & 86.10 & \textbf{73.08} & \textbf{86.47} \\
        \bottomrule
    \end{tabular}
\end{table}


\subsubsection{Loss Efficiency and Model Complexity Analysis}

Unlike methods such as InverseForm \cite{inverseForm}, which introduces an additional trainable boundary-transformation module, MABL requires no learnable parameters. Table~\ref{tab:loss_comparison} compares MABL with cross-entropy, the
SAM-based boundary loss\cite{SAM_RS}, and Active Boundary Loss (ABL)
\cite{wang2022active}. MABL achieves the highest mIoU and mAcc with only a marginal increase in training time over the baseline, while remaining faster than ABL. Although the SAM loss has a lower
reported training time, this measurement excludes the additional
preprocessing required to generate its SAM boundary priors. Overall,
MABL provides a favorable balance between boundary accuracy,
computational cost, and ease of integration.

Table~\ref{tab:model_complexity} highlights that BASeg secures the best segmentation accuracy while staying computationally practical. BASeg requires approximately $3.8\times$ fewer FLOPs and 66.6\% fewer parameters than PyramidMamba, and approximately $6.9\times$ fewer FLOPs and 19.4\% fewer parameters than Samba. 

\begin{table}[!t]
\centering
\footnotesize 
\setlength{\tabcolsep}{8pt} 
\caption{Comparison of boundary losses on Vaihingen with CMTFNet. Runtime is
measured per image.}
\label{tab:loss_comparison}
\begin{tabular}{lccc}
\toprule
Loss & Time (ms/image) & mIoU & mAcc \\
\hline
Cross-entropy & 9.36 & 69.24 & 80.47 \\
SAM loss      & 11.29 & 70.15 & 83.72 \\
ABL           & 18.91 & 70.72 & 83.25 \\
\hline
MABL & 18.27 & 71.27 & 86.34 \\
\bottomrule
\end{tabular}
\end{table}

\begin{table}[!t]
\centering
\footnotesize 
\setlength{\tabcolsep}{5pt}
\caption{Model complexity and segmentation performance on Vaihingen
at $256\times256$ resolution.}
\label{tab:model_complexity}
\begin{tabular}{lccc}
\toprule
Model & GFLOPs & Parameters (M) & mIoU \\
\hline
CMTFNet       & 8.57 & 30.07 & 69.24 \\
RS3Mamba      & 15.83 & 43.32  & 70.77 \\
PyramidMamba  & 64.70 & 125.09 & 69.63 \\
Samba         & 118.11 & 51.86 & 65.99 \\
\hline
Ours & 17.13 & 41.79 & 73.08 \\
\bottomrule
\end{tabular}
\end{table}

\subsection{Conclusion}
This work presents BASeg, a boundary-aware remote sensing semantic segmentation framework with Structural Penalties. It combines GSM and CFM with MABL, whose complementary Mahalanobis and angle penalties promote structurally coherent regions and sharply aligned object boundaries. We conduct comprehensive evaluations across diverse benchmarks, which demonstrates that BASeg achieves state-of-the-art performance. 
Ablation studies further validate the effectiveness of each component. These results demonstrate the superiority of BASeg in boundary segmentation across urban morphologies. We hope BASeg can advance development in urban analytics and environmental monitoring.

\bibliography{aaai2027}
\clearpage
\section{Supplemental Material}
\subsection{GCD-25k Dataset}
\textbf{Dataset composition}
The proposed \textbf{GCD-25k} dataset contains 25,000 satellite images collected from 10 cities across five continents, with pixel-wise annotations for three semantic classes: background, building, and road. To increase geographic diversity and capture variations in global urban morphology, the dataset includes cities from Asia (Singapore, Tokyo), Africa (Abuja, Kinshasa), the Americas (Boston, San Francisco, S\~ao Paulo), Europe (Barcelona, Paris), and Australia (Perth). Each city contributes 2,500 images, ensuring a balanced representation across locations.

\textbf{Image and annotation sources}
Satellite imagery is acquired from Mapbox, while building footprints and road network data are obtained from OpenStreetMap (OSM). The OSM vector layers are spatially aligned with the imagery and rasterized to generate pixel-level semantic masks. Each image patch has a resolution of $512 \times 512$ pixels and corresponds to a ground coverage of $400\,\text{m} \times 400\,\text{m}$.

\begin{figure*}[!t]  
    \centering
    \includegraphics[width=0.8\textwidth]{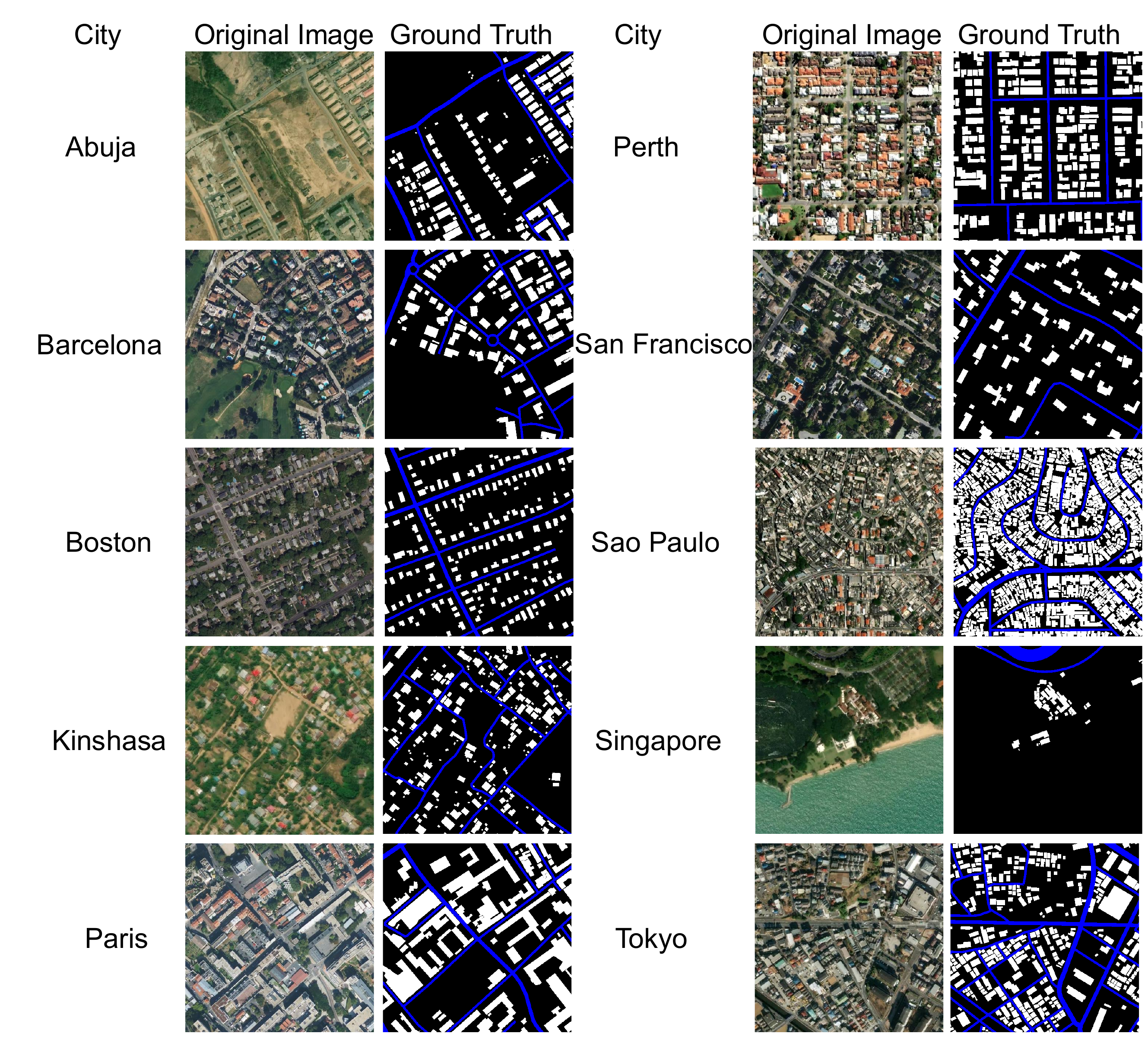} 
    \caption{Representative samples from the GCD-25k dataset across 10 cities, showing the original satellite image and the corresponding ground truth semantic mask.}
    \label{fig:GCD-25k}
\end{figure*}

\textbf{Sampling and split}
Images are sampled within the boundary of each city to provide broad spatial coverage and diverse urban layouts. The dataset is divided into two subsets: 20,000 training images and 5,000 testing images. For every city, 2,000 images are allocated to the training set and 500 images to the testing set, resulting in a balanced split across all cities. Sample images are shown in Figure \ref{fig:GCD-25k}, and additional city-level statistics are reported in Table \ref{tab:city_data_summary}.

\begin{table}[H]
\centering
\small
\caption{City-wise class coverage (\% of pixels).}
\label{tab:city_data_summary}
\setlength{\tabcolsep}{6pt}
\begin{tabular}{lcc}
\toprule
\textbf{City} & \textbf{Building (\%)} & \textbf{Road (\%)} \\
\midrule
Abuja         & 16.27 & 7.33 \\
Barcelona     & 23.04 & 9.89 \\
Boston        & 11.76 & 8.26 \\
Kinshasa      & 22.83 & 8.72 \\
Paris         & 22.69 & 9.19 \\
Perth         & 20.79 & 8.87 \\
San Francisco & 20.61 & 8.54 \\
Sao Paulo     & 35.84 & 11.58 \\
Singapore     & 23.15 & 9.10 \\
Tokyo         & 31.54 & 13.57 \\
\bottomrule
\end{tabular}
\end{table}

\subsection{Complete Per-Class Results}
To complement the quantitative results presented in the main paper, we report the complete per-class IoU scores for the HSRS and LoveDA datasets in Tables~\ref{tab:hsrs_iou} and~\ref{tab:loveDA-all}. Due to space limitations, the main paper reports only the overall evaluation metrics (mIoU and mAcc for HSRS, and mIoU for LoveDA). The detailed per-class results presented here provide a more comprehensive comparison across different semantic categories and further demonstrate the effectiveness of our proposed method. Following the convention used in the main paper, all IoU values in this section are reported as percentages.
\begin{table}[!t]
\centering
\scriptsize
\setlength{\tabcolsep}{3pt}
\caption{Per-class IoU comparison on the HSRS dataset. Bkg: Background; Bld: Building; Veg: Vegetation.}
\label{tab:hsrs_iou}
\begin{tabular}{lcccccccccc}
\toprule
Method &
Bkg &
Road &
Bld &
Veg &
Water &
Car &
Field &
Court &
Square &
mIoU \\
\midrule
ABCNet
& 76.23 & 56.13 & 58.16 & 69.38 & 74.50 & 69.38 & 50.08 & 77.70 & 30.73 & 62.48 \\

UNetformer
& 76.16 & 60.26 & 59.16 & 65.70 & 73.95 & 73.22 & 54.20 & 80.09 & 31.10 & 63.76 \\

CMTFNet
& 71.38 & 61.32 & 58.93 & 63.84 & 71.34 & \textbf{82.83} & 54.07 & 88.70 & 37.38 & 65.53 \\

RS3Mamba
& 78.24 & 67.25 & 64.01 & 69.53 & 77.53 & 78.73 & 56.14 & 84.54 & 37.73 & 68.19 \\

Samba
& 75.21 & 60.21 & 61.61 & \textbf{71.03} & 72.47 & 64.96 & 46.72 & 85.38 & 32.55 & 63.35 \\

PyramidMamba
& 77.40 & 67.24 & 62.73 & 69.87 & 76.38 & 47.26 & 57.36 & 87.30 & \textbf{43.11} & 65.41 \\
\hline
\textbf{Ours}
& \textbf{78.96}
& \textbf{67.43}
& \textbf{65.37}
& 70.21
& \textbf{78.10}
& 78.80
& \textbf{58.01}
& \textbf{89.21}
& 38.60
& \textbf{69.41} \\
\bottomrule
\end{tabular}
\end{table}

\begin{table}[!t]
    \centering
    \scriptsize
    \setlength{\tabcolsep}{4pt}

    \caption{Per-class IoU comparison of segmentation models on the LoveDA dataset. Bkg: Background; Bld: Building.}
    \label{tab:loveDA-all}

    \begin{tabular}{lccccccc}
        \toprule
        Method & Bkg & Bld & Road & Water & Forest & Agriculture & mIoU \\
        \midrule
        ABCNet        & 37.43 & 56.04 & 53.32 & 54.67 & 35.65 & 28.98 & 44.35 \\
        UNetformer    & 37.61 & 52.78 & 51.89 & 63.47 & 34.23 & 11.44 & 41.90 \\
        CMTFNet       & 30.71 & 46.29 & 53.24 & 46.61 & 38.65 & 28.17 & 40.61 \\
        Samba         & 34.20 & 38.74 & 43.11 & 62.70 & 35.87 & 17.11 & 38.62 \\
        PyramidMamba  & 35.02 & 58.06 & \textbf{59.82} & \textbf{63.74} & 31.31 & 18.48 & 44.44 \\
        \hline
        \textbf{Ours} & \textbf{36.93} & \textbf{59.19} & 56.87 & 62.72 & \textbf{39.82} & \textbf{29.62} & \textbf{47.53} \\
        \bottomrule
    \end{tabular}
\end{table}

\subsection{Loss Ablation on Remaining Datasets}
We further evaluate the contribution of the proposed Mahalanobis Distance Loss and Angle Loss on the LoveDA and HSRS datasets. Table~\ref{tab:loss-ablation-extra} reports the complete per-class IoU and accuracy results. Consistent with the observations on GCD-25k and Vaihingen in the main paper, both loss components contribute complementary improvements across different semantic categories, resulting in the best overall mIoU and mAcc when used jointly.
\begin{table}[!t]
    \centering
    \caption{Ablation study of loss functions on the LoveDA and HSRS datasets.}
    \label{tab:loss-ablation-extra}

    \scriptsize
    \setlength{\tabcolsep}{3.5pt}

    \begin{tabular}{c|l|cc|cc|cc}
        \toprule
        \multirow{2}{*}{Dataset} & \multirow{2}{*}{Class}
        & \multicolumn{2}{c|}{No Mahalanobis Loss}
        & \multicolumn{2}{c|}{No Angle Loss}
        & \multicolumn{2}{c}{Ours} \\
        \cline{3-8}
        & & IoU & Acc & IoU & Acc & IoU & Acc \\
        \hline

        \multirow{7}{*}{LoveDA}
        & Background  & 36.03 & 58.95 & 35.51 & 61.15 & \textbf{36.93} & \textbf{61.45} \\
        & Building    & 57.63 & 89.68 & 57.55 & 89.48 & \textbf{59.19} & 85.97 \\
        & Road        & 54.69 & 71.28 & 54.37 & 71.69 & \textbf{56.87} & 69.51 \\
        & Water       & 58.08 & 67.89 & 56.60 & 62.86 & \textbf{62.72} & \textbf{73.84} \\
        & Forest      & 42.52 & 85.41 & 34.99 & 84.01 & 39.82 & \textbf{87.05} \\
        & Agriculture & 26.09 & 27.95 & 25.06 & 27.13 & \textbf{29.62} & \textbf{30.73} \\
        \cline{2-8}
        & mIoU / mAcc & 45.84 & 66.86 & 44.01 & 66.05 & \textbf{47.53} & \textbf{68.09} \\
        \hline

        \multirow{10}{*}{HSRS}
        & Background  & 78.29 & 88.36 & 78.33 & 87.47 & \textbf{78.96} & \textbf{88.42} \\
        & Road        & 66.59 & 76.63 & 66.97 & 78.18 & \textbf{67.43} & \textbf{78.21} \\
        & Building    & 64.36 & 77.94 & 64.58 & 79.98 & \textbf{65.37} & \textbf{80.50} \\
        & Vegetation  & 69.25 & 81.59 & 69.91 & 82.02 & \textbf{70.21} & \textbf{82.16} \\
        & Water       & 76.91 & 87.79 & 77.49 & 87.56 & \textbf{78.10} & 87.49 \\
        & Car         & 78.35 & 88.19 & 77.00 & 83.15 & \textbf{78.80} & \textbf{88.21} \\
        & Field       & 58.56 & 73.35 & 57.17 & 72.69 & 58.01 & 70.73 \\
        & Court       & 82.07 & 98.33 & 84.29 & 97.62 & \textbf{89.21} & 97.37 \\
        & Square      & 37.84 & 52.67 & 37.54 & 50.90 & \textbf{38.60} & \textbf{53.84} \\
        \cline{2-8}
        & mIoU / mAcc & 68.02 & 80.54 & 68.14 & 79.95 & \textbf{69.41} & \textbf{80.77} \\
        \bottomrule
    \end{tabular}
\end{table}
\subsection{Detailed Ablation on Model Architecture}
On GCD-25k, we perform fine-grained component analysis. As shown in Table \ref{tab:gcd_gsm}, incorporating Coordinate Attention into the global branch leads to consistent gains in all three categories. Introducing the rectangular convolution module further benefits structure-aware classes, particularly buildings and roads, increasing their accuracy by 1.6\% and 0.8\%, respectively. To study the effect of kernel configuration, we compare two intermediate RCM variants: RCM-2, which employs two parallel rectangular convolutions ($3\times5$ and $5\times3$), and RCM-3, which additionally includes a $5\times5$ convolution. Based on these observations, we adopt a final RCM consisting of three parallel convolutions with kernel sizes $3\times5$, $3\times3$, and $5\times3$, which achieves the best overall performance. We then investigate the contribution of the Cross Fusion Module (CFM) in Table \ref{tab:gcd_cfm} and identified that introducing Spatial Attention improves overall IoU by 10\%. These results demonstrate the progressive benefits of each architectural component.
\begin{table}[!t]
    \centering
    \caption{Ablation study of GSM architecture on the GCD-25k dataset. RCM: Rectangular Convolution Module. (CFM is kept active in all configurations).}
    \label{tab:gcd_gsm}
    
    \scriptsize
    \setlength{\tabcolsep}{2pt}
    
    \begin{tabular}{l|cc|cc|cc|cc|cc}
        \toprule
        \multirow{2}{*}{Class}
        & \multicolumn{2}{c|}{No RCM}
        & \multicolumn{2}{c|}{No Coord. Attn.}
        & \multicolumn{2}{c|}{RCM-2}
        & \multicolumn{2}{c|}{RCM-3}
        & \multicolumn{2}{c}{GSM} \\
        \cline{2-11}
        & IoU & Acc & IoU & Acc & IoU & Acc & IoU & Acc & IoU & Acc \\
        \hline
        Building    & 63.38 & 76.43 & 61.32 & 75.84 & 63.97 & 77.89 & 64.22 & 77.01 & \textbf{63.87} & \textbf{78.00} \\
        Road        & 60.56 & 77.08 & 57.17 & 73.39 & 60.66 & 76.06 & 60.49 & 73.88 & \textbf{61.48} & \textbf{77.84} \\
        Background  & 81.10 & 89.74 & 76.62 & 86.73 & 78.46 & 87.90 & 79.07 & 89.09 & \textbf{81.24} & \textbf{89.36} \\
        \hline
        mIoU / mAcc & 68.34 & 81.08 & 65.04 & 78.65 & 67.70 & 80.62 & 67.93 & 79.99 & \textbf{68.86} & \textbf{81.73} \\
        \bottomrule
    \end{tabular}
\end{table}

\begin{table}[!t]
    \centering
    \caption{Ablation study of CFM architecture on the GCD-25k dataset. SA: Spatial Attention; CSA: Cross Attention. (GSM is kept active in all configurations).}
    \label{tab:gcd_cfm}
    
    \scriptsize 
    \setlength{\tabcolsep}{3.5pt} 
    
    \begin{tabular}{l|cc|cc|cc}
        \toprule
        \multirow{2}{*}{Class} 
        & \multicolumn{2}{c|}{No SA} 
        & \multicolumn{2}{c|}{No CSA} 
        & \multicolumn{2}{c}{CFM} \\
        \cline{2-7}
        & IoU & Acc & IoU & Acc & IoU & Acc \\
        \hline
        Building    & 54.81 & 68.66 & 59.12 & 72.75 & \textbf{63.87} & \textbf{78.00} \\
        Road        & 45.50 & 50.17 & 53.84 & 67.96 & \textbf{61.48} & \textbf{77.84} \\
        Background  & 76.13 & 87.76 & 79.05 & 89.26 & \textbf{81.24} & \textbf{89.36} \\
        \hline
        mIoU / mAcc & 58.82 & 72.20 & 64.00 & 76.66 & \textbf{68.86} & \textbf{81.73} \\
        \bottomrule
    \end{tabular}
\end{table}








\end{document}